\documentclass{bmvc2k}

\usepackage{times}
\usepackage{epsfig}
\usepackage{graphicx}
\usepackage{amsmath}
\usepackage{amssymb}

\usepackage{amsthm}
\usepackage{helvet}
\usepackage{courier}

\usepackage{url}
\usepackage{rotating}
\usepackage{xspace}
\usepackage{booktabs}       
\usepackage{nccmath}
\usepackage{amsfonts}       
\usepackage{nicefrac}       
\usepackage{microtype}      
\usepackage{comment}		
\usepackage{adjustbox}		
\usepackage{tabu}
\usepackage{calc}			
\usepackage{wrapfig}

\usepackage{multirow}
\usepackage{float}
\usepackage[hang,flushmargin]{footmisc}
\usepackage[neverdecrease]{paralist}
\usepackage{caption}
\usepackage{algorithm}
\usepackage{algorithmic}
\usepackage{calc}
\usepackage{color}
\usepackage{enumitem}
\setlist{leftmargin=4mm}

\newcommand{\calL}{{\mathcal{L}}}

\newcommand{\be}{\begin{eqnarray}}
\newcommand{\ee}{\end{eqnarray}}
\newcommand{\bee}{\begin{eqnarray*}}
\newcommand{\eee}{\end{eqnarray*}}

\newcommand{\matrixb}{\left[ \begin{array}}
\newcommand{\matrixe}{\end{array} \right]}

\renewcommand{\paragraph}[1]{\vspace{1mm}\noindent\textbf{#1}\quad}

\usepackage{colortbl}
\definecolor{CuGray}{gray}{0.9}
\newcolumntype{g}{>{\columncolor{CuGray}}c}

\makeatletter
\DeclareRobustCommand\onedot{\futurelet\@let@token\@onedot}
\def\@onedot{\ifx\@let@token.\else.\null\fi\xspace}

\newcommand{\dashrule}[1][black]{%
  \color{#1}\rule[\dimexpr.5ex-.2pt]{4pt}{.4pt}\xleaders\hbox{\rule{4pt}{0pt}\rule[\dimexpr.5ex-.2pt]{4pt}{.4pt}}\hfill\kern0pt%
}

\newcommand*{\Scale}[2][4]{\scalebox{#1}{$#2$}}%

\newcommand\blfootnote[1]{%
  \begingroup
  \renewcommand\thefootnote{}\footnote{#1}%
  \addtocounter{footnote}{-1}%
  \endgroup
}

\def\eg{\emph{e.g}\onedot} 
\def\ie{\emph{i.e}\onedot}

\def\etal{\emph{et al}\onedot}

\title{Visuomotor Understanding for Representation Learning of Driving Scenes}

\addauthor{Seokju Lee${}^{\dagger}$}{seokju91@gmail.com}{1}
\addauthor{Junsik Kim${}^1$\\[1mm]
Tae-Hyun Oh${}^2$\\[1mm]
Yongseop Jeong${}^1$\\[1mm]
Donggeun Yoo${}^3$\\[1mm]
Stephen Lin${}^4$\\[1mm]
In So Kweon}{}{1} %

\addinstitution{
 KAIST\\
 Daejeon, Korea
}
\addinstitution{
 MIT CSAIL\\
 Cambridge, MA, USA
}
\addinstitution{
 Lunit Inc.\\
 Seoul, Korea
}
\addinstitution{
 Microsoft Research Asia\\
 Beijing, China
}

\runninghead{Lee et al.}{Visuomotor Understanding for Representation Learning}

\def\eg{\emph{e.g}\bmvaOneDot}

\def\etal{\emph{et al}\bmvaOneDot}

\begin{document}

\maketitle

\blfootnote{$^\dagger$Part of this work was done while S. Lee was at Microsoft Research Asia.}

\begin{abstract}
Dashboard cameras capture a tremendous amount of driving scene video each day. These videos are purposefully coupled with vehicle sensing data, such as from the speedometer and inertial sensors, providing an additional sensing modality for free. In this work, we leverage the large-scale unlabeled yet naturally paired data for visual representation learning in the driving scenario. A representation is learned in an end-to-end self-supervised framework for predicting dense optical flow from a single frame with paired sensing data. We postulate that success on this task requires the network to learn semantic and geometric knowledge in the ego-centric view. For example, forecasting a future view to be seen from a moving vehicle requires an understanding of scene depth, scale, and movement of objects. We demonstrate that our learned representation can benefit other tasks that require detailed scene understanding and outperforms competing unsupervised representations on semantic segmentation. 
\end{abstract}

\vspace{-2mm}
\section{Introduction}
\vspace{-2mm}



An essential capability for intelligent vehicles is understanding causal relationships between its motion and the surrounding environment. Knowing how its movement affects what it would see around it can aid the vehicle in selecting safe and proper courses of action.


The ability to synchronize visual information with physical movement is commonly referred to as \emph{visuomotor understanding}. For humans, this understanding is critical for daily functioning, as 80\% of human perception depends on vision, and most sensory decision-making is aimed toward movement~\cite{goldstein2016sensation}. This coordination of fine motor skills with visual stimuli is developed from infancy with basic movements such as toddling and eventually improves to perform more complex tasks like buttoning shirts and tying shoelaces~\cite{jakobson2001relationship}.


Motivated by the human perception system, we develop an unsupervised framework for developing visuomotor understanding in driving scenes from paired visual and ego-motion sensory information. 
One of our main goals is to \textit{learn a visual representation} by predicting future frames via dense motion fields from fused visual and ego-motion data. We argue that for effective inference in this task, the model needs to learn semantic and geometric knowledge with respect to the ego-centric viewpoint. Specifically, forecasting future frame appearance driven by motion requires comprehensive understanding of scene depth, object scale, and movements of dynamic objects.


Towards this goal, we propose a novel deep network that takes as input a single frame together with the corresponding motion sensor data, and estimates dense optical flow for predicting the appearance of the next frame. The motion sensor data is concatenated with the encoded visual features after undergoing a learned embedding into a latent space. The predicted flow is used to warp the input frame forward by one-time step, and the training loss is defined based on the difference between the warped image and the actual next frame.
A key property of the proposed method is its \emph{time reversal symmetry (T-symmetry)}~\cite{sachs1987physics}.
Our work takes the physical variables of \emph{velocity} and \emph{angular momentum} which are affected by time reversal that can be used as control inputs in the network and also to introduce additional self-supervision as described in Sec.~\ref{sec:3_4}.
For training, we have collected large-scale pairs of image and motion data by simply driving a vehicle equipped with a camera and a mobile sensor that measures global speed and inertia.
After large-scale training with the proposed framework, we finetune our model on a semantic segmentation task with a public dataset to verify its transferability.

\vspace{1mm}
\noindent\textbf{Contributions}\quad
 To sum up, the main contributions of this work are as follows.
\vspace{-2mm}
\begin{enumerate}
\item~A generic sensor fusion architecture that predicts dense optical flow for synthesizing future or past frames with the help of motion sensor data and time reversal symmetry. The effectiveness of these components is validated by extensive ablation studies.
\vspace{-2mm}
\item~A visual representation learned by our method is shown to be effective for semantic segmentation in the autonomous driving scenario and useful for other vision applications.
\end{enumerate}

\vspace{-2mm}
\section{Related Works}
\vspace{-2mm}

\paragraph{Visual representation learning}
Many previous works~\cite{doersch2015unsupervised,noroozi2016unsupervised,pathak2016context,zhang2016colorful,noroozi2016unsupervised,noroozi2017representation,larsson2017colorization} for unsupervised visual representation learning have aimed to acquire high-level understanding within a single-image context.
Beyond the scope of a single image, several recent works have leveraged an additional dimension of data, such as temporal sequences~\cite{wang2015unsupervised,misra2016shuffle,zhou2017unsupervised,pathak2017learning,jiang2018self} and multi-modal input~\cite{agrawal2015learning,jayaraman2015learning}.
Our work lies in the direction of multi-modal based representation learning, specifically utilizing motor sensor and visual information in a collaborative fashion.


Learning general visual representations from multi-modal data has been addressed in the context of driving scenes~\cite{agrawal2015learning,jayaraman2015learning}. 
Agrawal~\etal~\cite{agrawal2015learning} learn a representation for predicting the camera transformation between a pair of input images, with recorded ego-motion as self-supervision. 
Given pairs of images and the direction of motion between them, Jayaraman and Grauman~\cite{jayaraman2015learning} acquire an equivariant representation, where the relative positions in the feature space of two images can be predicted by the motion direction between them. 
Compared to these methods, our work aims to learn a representation with stronger knowledge of scene structure. 
As Sax~\etal~\cite{sax2018mid} studied, robotic locomotive tasks,~\eg, visual exploration or navigation, require understanding of mid-level visual features~\cite{peirce2015understanding}.
Learning to predict the change in viewed scene appearance with respect to ego-motion requires more detailed understanding of scene geometry, including occlusions and disocclusions from camera motion, than what is needed to estimate camera pose change between a pair of images~\cite{agrawal2015learning} or relative feature space displacements~\cite{jayaraman2015learning}.
We demonstrate that our learned representation is more effective than these approaches on important driving-related tasks that benefit from structural scene understanding, such as semantic segmentation.

\paragraph{Learning view synthesis}
Visuomotor understanding involves the ability to predict changes in frame appearance that accompany camera motion. We use this view synthesis problem as a \emph{proxy task} for learning a visual representation. In other works on view synthesis, 
Kulkarni~\etal~\cite{kulkarni2015deep} and Yang~\etal~\cite{yang2015weakly} disentangled latent pose factors of an image, limited to rotations of simple objects such as faces or chairs.
View interpolation~\cite{flynn2016deepstereo,ji2017deep} and extrapolation~\cite{zhou2018stereo} methods synthesize high-quality novel views, but require more than two input frames.
Tatarchenko~\etal~\cite{tatarchenko2016multi} proposed an encoder-decoder network to directly regress the pixels of a new image from a single input image, but tends to produce blurry results.
Zhou~\etal~\cite{zhou2016view} alleviated this problem through a flow-based sampling approach called \textit{appearance flow}, but this often generates artifacts due to warped scene structure.
Recently, Liu~\etal~\cite{liu2018geometry} exploited 3D geometry to synthesize a novel view using depth labels, and Park~\etal~\cite{park2017transformation} and Sun~\etal~\cite{sun2018multi} jointly trained flow-based pixel generation networks, but these works are geared toward a specific application, rather than learning a visual representation that can be used for various semantic understanding tasks.

\paragraph{Learning optical flow}
Estimating optical flow formally requires at least two input images.
Although Pintea~\etal\cite{pintea2014deja} proposed a method for single-image flow prediction, they dealt only with human actions.
Obtaining optical flow between two images is a well-studied computer vision problem~\cite{revaud2015epicflow,brox2011large,sun2014quantitative}.
Several recent works have proposed CNN-based supervised learning methods~\cite{weinzaepfel2013deepflow,dosovitskiy2015flownet,mayer2016large,ilg2017flownet,sun2017pwc} with ground truth flow, and unsupervised learning methods~\cite{jason2016back,ahmadi2016unsupervised,ren2017unsupervised,meister2017unflow} with unlabeled pairs of images.
However, these approaches for optical flow are not suitable for learning a general semantic representation, because they focus on learning to match local areas between two images, which does not require holistic scene understanding and semantic knowledge.

This difference between our work and existing flow estimation methods can be further explained as follows.
Flow estimation with two sequential images, $I_t$ and $I_{t+1}$, is formulated as
$F_{t, t+1}=\mathcal{F}(I_t,\;I_{t+1})$,
where $\mathcal{F}$ is a conventional model for estimating optical flow.
By contrast, our newly proposed flow prediction method with motion sensor modality, $S_t$, can be represented as
$F_{t, t+1}=\widetilde{\mathcal{F}}(I_t,\;S_t$),
where $\widetilde{\mathcal{F}}$ is our model, called SensorFlow.
While the function $\mathcal{F}(\cdot)$
is learned from how to match the two images photometrically, our function $\widetilde{\mathcal{F}}(\cdot)$ does not learn such a comparator, as only a single image is given.
By learning our function with respect to a static scene image and a physical motion $S_t$, it is forced to learn a representation based on structural and semantic understanding, rather than a representation targeted at local matching.

\vspace{-2mm}
\section{SensorFlow Architecture}
\vspace{-2mm}

Our objective is to train a non-linear mapping to predict optical flow given an RGB image and synchronized sensor data. In this section, we introduce the SensorFlow architecture to achieve this goal. Given an RGB image, the network estimates optical flow, of which the direction is controlled by the input sensor data.
Further, we describe how sensor values are used as control parameters and fused with the visual representation, and explain the loss functions used to train the network.

\begin{figure*}[t]
\centering
\includegraphics[width=0.97\textwidth]{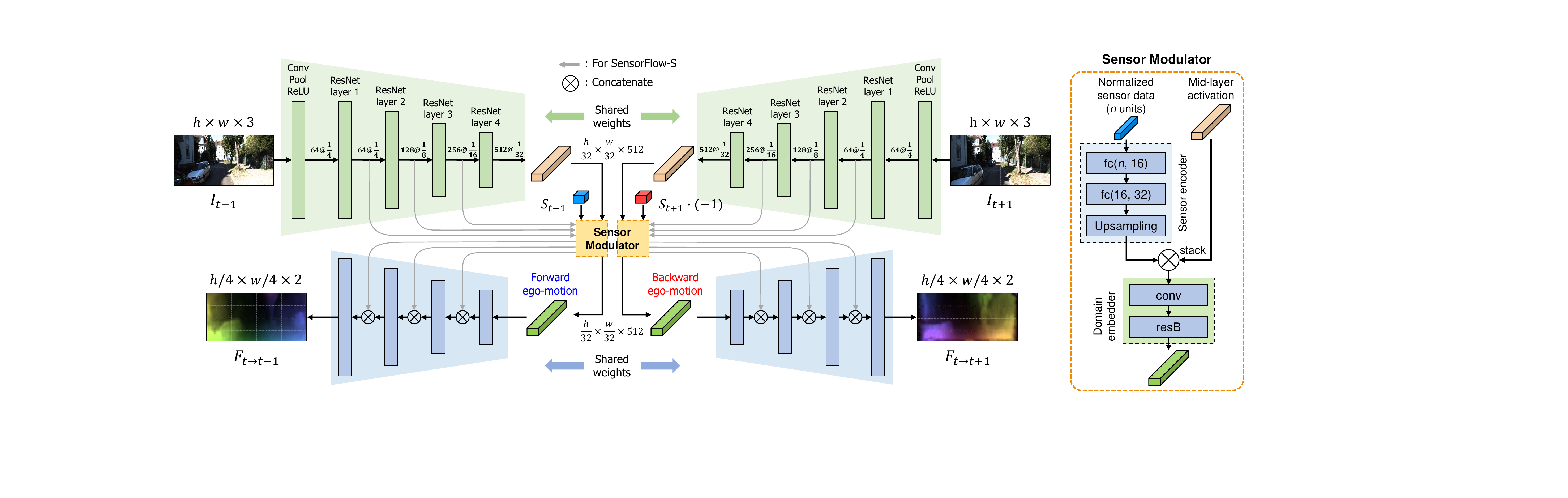}
\vspace{+3mm}
\caption{
Illustration of SensorFlow architecture and its sensor modulator. The base encoder here is a ResNet. The network is trained on image data ($I$) and sensor data ($S$) collected from a vehicle while driving. The sensor modulator controls the direction of the flow by encoding the sensor data into the visual domain (fc: fully-connected layer, conv: convolutional layer, resB: residual block). A natural causal relationship exists between this vehicle data and flow fields ($F$). Leveraging this relationship, our network learns to predict the current frame ($\hat{I}_t$) from a past frame ($I_{t-1}$) or a future frame ($I_{t+1}$) given sensor data that is embedded in the latent space. By increasing its visuomotor understanding in this manner, our network learns a visual representation built on semantic and geometric knowledge of driving scenes.
}
\vspace{-3mm}
\label{fig:architecture}
\end{figure*}

\vspace{-2mm}
\subsection{Basic Architecture}
\label{sec:3_1}
\vspace{-2mm}

We designed a simple and novel network that utilizes motion sensor information to learn the relationship between ego-motion and changes in scene appearance while learning a versatile visual representation.
The basic architecture of SensorFlow is illustrated in Figure~\ref{fig:architecture}.

Let us first focus on the left side of the architecture in Figure~\ref{fig:architecture}.
SensorFlow consists of an encoding part to extract visual features from an image and a decoding part to decode the features into optical flow. Our ultimate goal of training SensorFlow is to obtain an encoder network which can be reused for various recognition tasks such as semantic segmentation in driving scenes.
To this end, we design the encoder to be compatible with any general-purpose network architecture such as AlexNet~\cite{krizhevsky2012imagenet}, VGG~\cite{simonyan2014very} or ResNet~\cite{witten2016data}.

For the decoding part, we stack multiple deconvolution layers to upsample the feature map as done in~\cite{dosovitskiy2015flownet}. We also employ skip connections as in~\cite{ronneberger2015u} where the intermediate feature maps of the encoder are passed to the decoder to enhance fine detail in the output. To accommodate various backbone architectures, the skip connections are applied in a layer-symmetric manner. This version of SensorFlow containing these skip connections is denoted as SensorFlow-S.

Let us take a look at both the left and right sides of the architecture in Figure~\ref{fig:architecture}.
To implement the idea of \emph{T-symmetry}, we design the network as two streams so that it can learn to predict bidirectional flows simultaneously for the forward and backward motions. Specifically, given three temporally consecutive images $I_{1}$, $I_{2}$ and $I_{3}$, the left side of the network generates a flow map $F_{2\rightarrow 1}$ from $I_{1}$, and the right side of the network generates a flow map $F_{2\rightarrow 3}$ from $I_{3}$. Details on this are given in the following sections.
For further details on the architecture, readers can refer to the supplementary material.

\vspace{-2mm}
\subsection{Learning Sensor Representations}
\label{sec:3_2}
\vspace{-2mm}
Predicting optical flow or a neighboring frame from a single image is an ill-posed problem. However, with information about camera motion under an assumption that the surrounding environment is static, we can predict the global flow of the scene structure. Here, our goal is to estimate a fine flow map from only a single image frame and paired sensor values, with performance comparable to two-view flow methods.

Given a single image $I_t:X\rightarrow \mathbb{R}^3$ at time $t$, we define forward and backward sensor data $S^+_t$ and $S^-_t$ as follows:
\begin{align}
\Scale[0.97]{
S^+_t=\left[ s_1 \; s_2 \; s_3 \; \cdots \; s_n \right]^\top, \;\;
S^-_t=-\left[ s_1 \; s_2 \; s_3 \; \cdots \; s_n \right]^\top,
}
\end{align}
where $n$ denotes the number of sensor measurements.
Given the inputs $I_t$ and $S_t$, the predicted flow maps, $F$, and generated image frames, $\hat{I}$, of both forward and backward motions are represented as follows:
\begin{align}
\Scale[0.97]{
F_{t+1\rightarrow t}=f_\text{flow}(I_t,\;S^+_t), \;\;
F_{t-1\rightarrow t}=f_\text{flow}(I_t,\;S^-_t),
}
\end{align}
\vspace{-6mm}
\begin{align}
\Scale[0.93]{
\hat{I}^{f}_{t+1}=f_\text{warp}(I_t,\;F_{t+1\rightarrow t}), \;\;
\hat{I}^{b}_{t-1}=f_\text{warp}(I_t,\;F_{t-1\rightarrow t}),
}
\end{align}
where $f_\text{flow}$ is the function for flow prediction and $f_\text{warp}$ is the function for image warping using a differentiable grid sampling layer proposed by Jaderberg~\etal~\cite{jaderberg2015spatial}.
Note that the grid sampling layer is used to transform an image in the reverse direction of the flow.

\vspace{-2mm}
\subsection{Sensor Modulator}
\label{sec:3_3}
\vspace{-2mm}

A key element of SensorFlow is a proposed sensor modulator that can encode a vector of sensor signals into the visual domain. The sensor modulator receives two inputs, normalized sensor data and a mid-layer activation. For sensor data preprocessing, we perform normalization by obtaining the mean and standard deviation over the entire training set for each of the $n$ sensor units. At each time step, both forward and backward data are processed concurrently in training, and the average sensor value for each unit between the two directions is zero, even after normalization. As discussed later for T-symmetry, this property will be utilized for regularization. Another input, a mid-layer activation, is the neural output from an encoding layer. For the basic SensorFlow model without the skip-connection structure, this is the final output, which is a latent variable of the encoder.

The sensor modulator is divided into two parts: a sensor encoder and a domain embedder. Figure~\ref{fig:architecture} shows the structure of our sensor modulator.
First, the sensor encoder transforms the sensor values into the visual domain. This is done via two fully-connected layers that extend the channel size, and an upsampling layer that expands the spatial size to the same resolution as the mid-layer activations. This expansion is achieved by repeating the same $1\times 1$ vector to a size of $h\times w$. In SensorFlow-S, the weights of the sensor encoder are shared for all mid-layer activations.
Second, the domain embedder stacks the sensor feature plane with the mid-layer activation plane and converts them into a common domain via one convolutional layer and one residual block~\cite{he2016deep}.
As a design note, the sensor modulator does not include any normalization layer (\emph{e.g.} batch normalization~\cite{ioffe2015batch}, local response normalization), as the neurons must preserve the scale of the motions.
Each convolutional layer is followed by a ReLU.
The generated encoding contains visual information as well as information on the direction and scale of the motion.

\vspace{-2mm}
\subsection{Self-Supervised Loss}
\label{sec:3_4}
\vspace{-2mm}

Similar to the loss in~\cite{jason2016back}, we use an unsupervised loss that measures the photometric inconsistency between $I$ and $\hat{I}$.
Since the photometric loss does not reflect the movement of dynamic objects or non-rigid motions, we apply the structural similarity index SSIM~\cite{wang2004image} to mitigate the effects of this movement. Our basic image warping cost with forward motion is written as
\begin{equation}
\Scale[0.9]
{
\begin{aligned}
\calL_{w}({\hat I}^{f}, \mathbf{M}^{f}) = 
\sum\limits_{{\rm{x}} \in X}  \left\{ {{\lambda _{1}} \rho \left( {{\bf{M}}^f}({\rm{x}}) \cdot \left\| { {I({\rm{x}}) - {{\hat I}^f}({\rm{x}})} } \right\|_{1} \right) } 
+ {\lambda _{2}} \left( 1 - SSIM(I({\rm{x}}), {{\hat I}^f}({\rm{x}})) \right)  \right\} ,
\end{aligned}
}
\end{equation}
where $\textrm{x}$ indicates each pixel location and $\rho(x)=(x^2+\epsilon^2)^\alpha$ is the robust generalized Charbonnier penalty function~\cite{sun2014quantitative} with $\alpha=0.4$. This function is equal to the original Charbonnier penalty when $\alpha=0.5$, which is a differentiable variant of the absolute function.
$\lambda _{1}$ and $\lambda _{2}$ are set to 0.3 and 0.7 respectively.
In order to exclude invalid gradients from occluded or exiting regions, we follow \cite{meister2017unflow} by setting the forward valid mask ${\bf{M}}^{f}(\rm{x})$ to be 1 if the condition
\begin{equation}
\Scale[0.94]
{
\begin{aligned}
{\left| {F^f({\rm{x}}) + F^b\left( {{\rm{x}}+F^f({\rm{x}})} \right)} \right|^2} 
> {\gamma _1}\left( {{{\left| {F^f({\rm{x}})} \right|}^2} + {{\left| {F^b\left( {{\rm{x}} + F^f({\rm{x}})} \right)} \right|}^2}} \right) + {\gamma _2},
\end{aligned}
}
\end{equation}
is satisfied, and 0 otherwise. We set $\gamma_1=0.01$, $\gamma_2=0.5$. For the backward valid mask ${\bf{M}}^{b}(\rm{x})$, we exchange $F^f$ and $F^b$ in the above condition. Each forward and backward flow, $F^f$ and $F^b$, is processed on two consecutive frames, \ie, ${\bf{M}}^{f}(\rm{x})$ by $\{I_{t-1}, I_{t}\}$ and ${\bf{M}}^{b}(\rm{x})$ by $\{I_{t+1}, I_{t}\}$.

To regularize the bidirectional training, we design a forward and backward flow consistency check in our learning scheme. This consistency check is based on the observation that within a short time interval, the flow of rigid objects generated by camera ego-motion can be linearly modeled~\cite{im2015high}, such that incremental flows in the forward and backward directions should sum to zero. Previous works~\cite{godard2017unsupervised,yin2018geonet} utilized a related idea in their depth prediction frameworks with a geometric consistency loss. We exclude both forward and backward occluded regions from the consistency check. Specifically, our bidirectional flow consistency cost is imposed as
\begin{equation} \label{eq_6}
\Scale[0.98]
{
\begin{aligned}
\calL_{c} (F_{t \to t - 1}^{f}, F_{t \to t + 1}^{b}, \mathbf{M}^{f}, \mathbf{M}^{b}) = 
\sum\limits_{{\rm{x}} \in X} {{{\bf{M}}^f}({\rm{x)}} \cdot {{\bf{M}}^b}({\rm{x)}} \cdot \left( {F_{t \to t - 1}^f({\rm{x}}) + F_{t \to t + 1}^b({\rm{x}})} \right)} ,
\end{aligned}
}
\end{equation}
where each non-occluded pixel $\rm{x}$ is enforced to have consistent flow magnitudes between its bidirectional motions.

As done in previous methods~\cite{dosovitskiy2015flownet,jason2016back}, we adopt a smoothness cost, $\calL_{s}$. The smoothness term is used to suppress spatial fluctuations.
We have empirically found that a relatively small loss weight for the smoothness term improves flow prediction.

To sum up, our final self-supervised loss is defined as
\begin{equation}
\Scale[0.95]
{
\begin{aligned}
\calL_{tot} = \lambda_{w} \left( \calL_{w}({\hat I}^{f}, \mathbf{M}^{f}) + \calL_{w}({\hat I}^{b}, \mathbf{M}^{b}) \right)
+ \lambda_{s} \left( \calL_{s}(F_{t \to t - 1}^{f}) + \calL_{s}(F_{t \to t + 1}^{b}) \right) \\
+ \lambda_{c} \calL_{c} (F_{t \to t - 1}^{f}, F_{t \to t + 1}^{b}, \mathbf{M}^{f}, \mathbf{M}^{b}) ,
\end{aligned}
}
\end{equation}
where $\lambda$ denotes loss weights. We set $\lambda_{w}=\lambda_{c}=1$ and $\lambda_{s}=0.1$. The total loss is measured in a bidirectional manner with warped forward and backward images.

\vspace{-1mm}
\section{Experiments}

\vspace{-1mm}
\subsection{Training}
\vspace{-1mm}

\paragraph{Our dataset}
For the representation learning of driving scenes, we collected a large-scale set of paired image and motion data from driving a vehicle equipped with a camera and a mobile sensor that measures global speed and various inertial quantities. Nearly 350,000 frames were obtained at 10~Hz from 12 cities and 11 countryside routes by driving 757 $km$ under various climate conditions. Detailed comparisons with existing driving datasets~\cite{geiger2012we,cordts2016cityscapes,maddern20171,santana2016learning,xu2017end,udacity2017public,ramanishka2018toward,chen2018lidar} and necessity of ours are presented in the supplementary material.

\paragraph{Proxy task}
For experiments involving the proxy task, including the ablation study and view synthesis experiments, training is done using the KITTI dataset, as it provides ground truth optical flow for quantitative evaluation.
The network is trained by the ADAM optimizer~\cite{kingma2014adam} for 350K iterations with a batch size of 20 on an Nvidia Titan X GPU and
an Intel i7@3.4GHz CPU. The initial learning rate is set to 0.0002, and it is decreased by half every 100K iterations.
While training, we take three consecutive frames as input to our two-stream network.
The observed sensor set of each frame is $\{v_x, v_y, v_z, \omega_x, \omega_y, \omega_z\}$, where $v_x$ and $\omega_x$ are the linear velocity and angular velocity along the $x$ axis.
Note that the sampling time, $\Delta{t}$, is different for each dataset (\eg~Ours and KITTI: $\Delta{t}=100~ms$, Cityscapes: $\Delta{t}\simeq 60~ms$).
Also, we average the sensor readings of three consecutive frames to reduce noise in the training data.


\paragraph{Representation learning task}
For experiments on representation learning, we pretrain our models using our large-scale dataset, and finetuned on the CamVid and CityScapes datasets for various architectures, namely the original AlexNet, VGG16, ResNet18, and ResNet34, using the same training techniques as in their respective works~\cite{krizhevsky2012imagenet,simonyan2014very,he2016deep}. We start the finetuning with a learning rate of 0.0001.

The models are evaluated on the Cityscapes~\cite{cordts2016cityscapes} and CamVid~\cite{brostow2009semantic} datasets.
Specifically, the evaluation uses the Cityscapes training set (3,000 images) and validation set (500 images), as well as the CamVid training set (367 images) and test set (233 images).
The Cityscapes dataset contains high resolution images which requires large GPU memory when training deep networks, so we downsize these images by half for training and evaluation. It is reported that downscaled images have consistently negative effects on both training and test~\cite{cordts2016cityscapes}. The gap between accuracy values found in our experiments and those previously reported in other works is mainly due to the image size difference.

\begin{table*}[t]
\caption{
SensorFlow ablations on KITTI 2012 optical flow dataset. Photometric error is averaged over forward and backward view syntheses, and EPE is averaged endpoint error. 
}
\label{tab_ablation}
\vspace{+3mm}
\centering
\huge
\begin{adjustbox}{width=0.81\textwidth}
\begin{tabular}{@{}cc|gcgcgcgcg|cgc|gcg@{}}
\toprule
\multicolumn{2}{c|}{ \multirow{2}{*}{Options} }
& \multicolumn{9}{c|}{Trials} & \multicolumn{3}{c|}{SensorFlow} & \multicolumn{3}{c}{SensorFlow-S} \\
\multicolumn{2}{c|}{}& $1^{st}$ & $2^{nd}$ & $3^{rd}$ & $4^{th}$ & $5^{th}$ & $6^{th}$ & $7^{th}$ & $8^{th}$ & $9^{th}$ & $1^{st}$ & $2^{nd}$ & $3^{rd}$ & $1^{st}$ & $2^{nd}$ & $3^{rd}$ \\
\midrule
\multirow{6}{*}{ \rotatebox{90}{\textit{Training}} }
& \multicolumn{1}{|c|}{Sensor modality} & & \checkmark & \checkmark & \checkmark & \checkmark & \checkmark & \checkmark & \checkmark & \checkmark & \checkmark & \checkmark & \checkmark & \checkmark & \checkmark & \checkmark \\
& \multicolumn{1}{|c|}{Bidirectional motion} & \checkmark & & \checkmark & \checkmark & \checkmark & \checkmark & \checkmark & \checkmark & \checkmark & \checkmark & \checkmark & \checkmark & \checkmark & \checkmark & \checkmark \\
& \multicolumn{1}{|c|}{Flow consistency} & & & & \checkmark & \checkmark & \checkmark & \checkmark & \checkmark & \checkmark & \checkmark & \checkmark & \checkmark & \checkmark & \checkmark & \checkmark \\
& \multicolumn{1}{|c|}{Skip-connection} & & & & & \checkmark & & & & & & & & \checkmark & \checkmark & \checkmark \\
& \multicolumn{1}{|c|}{Horizontal flip} & & & & & & \checkmark & \checkmark & \checkmark & \checkmark & \checkmark & \checkmark & \checkmark & \checkmark & \checkmark & \checkmark \\
& \multicolumn{1}{|c|}{Time variation} & & & & & & & \checkmark & \checkmark & \checkmark & \checkmark & \checkmark & \checkmark & \checkmark & \checkmark & \checkmark \\
\midrule
\multirow{4}{*}{ \rotatebox{90}{\textit{Modulator}} }
& \multicolumn{1}{|c|}{stack} & & \checkmark & \checkmark & \checkmark & \checkmark & \checkmark & \checkmark & & & & & & & & \\
& \multicolumn{1}{|c|}{stack+conv} & & & & & & & & \checkmark & & & & & & & \\
& \multicolumn{1}{|c|}{stack+conv+resB} & & & & & & & & & \checkmark & & & & & & \\
& \multicolumn{1}{|c|}{fc(2)+stack+conv+resB} & & & & & & & & & & \checkmark & \checkmark & \checkmark & \checkmark & \checkmark & \checkmark \\
\midrule
\multirow{3}{*}{ \rotatebox{90}{\textit{Units}} }
& \multicolumn{1}{|c|}{$v_x, v_y, v_z, w_x, w_y, w_z$} & & \checkmark & \checkmark & \checkmark & \checkmark & \checkmark & \checkmark & \checkmark & \checkmark & \checkmark & & & \checkmark & & \\
& \multicolumn{1}{|c|}{$v_x, w_x, w_y, w_z$} & & & & & & & & & & & \checkmark & & & \checkmark & \\
& \multicolumn{1}{|c|}{$v_x, w_z$} & & & & & & & & & & & & \checkmark & & & \checkmark \\
\midrule
\multicolumn{2}{c|}{Photometric error} & \Large 0.340 & \Large 0.269 & \Large 0.207 & \Large 0.194 & \Large 0.192 & \Large 0.193 & \Large 0.190 & \Large 0.189 & \Large 0.186 & \Large 0.184 & \Large 0.186 & \Large 0.201 & \Large 0.183 & \Large 0.185 & \Large 0.199 \\
\multicolumn{2}{c|}{EPE} & \Large 24.22 & \Large 16.70 & \Large 15.39 & \Large 14.91 & \Large 14.18 & \Large 14.76 & \Large 14.05 & \Large 14.02 & \Large 13.80 & \Large 13.77 & \Large 13.79 & \Large 15.11 & \Large 13.35 & \Large 13.68 & \Large 14.93 \\
\bottomrule
\end{tabular}
\end{adjustbox}
\vspace{-3mm}
\end{table*}

\vspace{-1mm}
\subsection{Ablation Study}
\vspace{-1mm}

\paragraph{Design process}~The ultimate goal of this work is to learn a visual representation for the driving scenario through the estimation of neighboring frames. In this section, we conduct an ablation study to verify that this is accurately estimated by SensorFlow. This study comprises three parts as shown in Table~\ref{tab_ablation}.
The first part considers training options. The second part is on how to embed the sensor readings. Finally, we compare the performance for different sensor combinations in a driving environment.
All ablation experiments are conducted by training ResNet18-based SensorFlow models on the KITTI raw dataset. The performances are compared using the average photometric error of forward and backward warping and the average endpoint error (EPE) on the KITTI 2012 optical flow dataset.

\paragraph{Regularization}~To verify the effect of bidirectional training based on \emph{T-symmetry}, models trained with only a forward motion, and with both forward and backward motions are compared ($2^{nd}$ and $3^{rd}$ columns of the trials in Table~\ref{tab_ablation}). It was found that the model without bidirectional motion is easily biased to always predict flow with forward motion, regardless of the sensor readings.
Another advantage of bidirectional training comes from the flow consistency loss as proposed in Equation~\ref{eq_6}. Ablations without and with this loss ($3^{rd}$ and $4^{th}$ columns of the trials in Table~\ref{tab_ablation}) show that our bidirectional flow consistency term improves performance considerably via constraints on the opposite flow directions.

We utilize two forms of data augmentation for regularization. One is the common technique of image flipping, which yields improvements from comparison of the $4^{th}$ and $6^{th}$ columns of the trials in Table~\ref{tab_ablation}. The other is to vary the time intervals of optical flows, \eg, by also generating $I_3$ from $I_1$ and $I_1$ from $I_3$ with $2\cdot S^+$ and $2\cdot S^-$, respectively. This leads to a significant improvement from the $6^{th}$ and $7^{th}$ columns of the trials in Table~\ref{tab_ablation}.

More descriptions on other design choices, \eg, sensor embedding and controllability, are presented in our supplementary material.

\subsection{View Synthesis}
To demonstrate that the proxy task is effectively learned, we conduct
experiments on view synthesis.
We control to the sensory input to synthesize a new view from a different viewpoint. 
Table~\ref{tab_synthesis} shows that our proposed method performs favorably against the competing appearance flow techniques while accounting for the number of parameters of each model.
Detailed experimental settings are given in the supplementary material due to limited space.
The results indicate the effectiveness of embedding the control variables from the sensor into the continuous latent space.
Note that the purpose of view synthesis is to validate whether our representations are plausibly learned to understand scene changes according to sensor inputs, rather than to generate visually pleasing results.

Furthermore, we qualitatively test our network by generating a new view and applying a stereo matching algorithm between an input and its new synthesized view, \ie, single view depth estimation. This allows us to see whether our network learns plausible depth perception capability. 
As shown in the supplementary material, the results indicate that our model is potentially extensible to single-view depth estimation.

\begin{table*}
\centering
\begin{minipage}{.49\textwidth}
\renewcommand{\tabcolsep}{5mm}
\vspace{-1.5mm}
\caption{Photometric errors of view synthesis on KITTI with different time steps.}
\label{tab_synthesis}
\vspace{+11.5mm}
\begin{adjustbox}{width=1\textwidth}
\begin{tabular}{lccc}
\toprule
Method & Parameters & \begin{tabular}[c]{@{}c@{}}$\pm$ One\\ time step\end{tabular} & \begin{tabular}[c]{@{}c@{}}$\pm$ Two\\ time step\end{tabular} \\
\midrule
MV3D~\cite{tatarchenko2016multi} & 69.3M & 0.241  & 0.316 \\
Appearance Flow~\cite{zhou2016view} & 5.5M & 0.223  & 0.285 \\
\midrule
SensorFlow (AlexNet) & 4.6M & 0.191 & 0.239 \\ 
SensorFlow (ResNet34) & 29.2M & 0.178 & 0.212 \\ 
SensorFlow-S (ResNet34) & 31.3M & 0.173 & 0.204 \\ 
\bottomrule
\end{tabular}
\end{adjustbox}
\end{minipage}%
\hspace{4mm}
\begin{minipage}{.46\textwidth}
\renewcommand{\tabcolsep}{1mm}
\caption{Mean IoU comparisons for semantic segmentation.}
\vspace{1.5mm}
\label{tab_segmenation}
\begin{adjustbox}{width=1\textwidth}
\begin{tabular}{@{}cgcgcgc@{}}
\toprule
Dataset & \multicolumn{2}{c}{CamVid} & \multicolumn{4}{c}{Cityscapes} \\
\cmidrule(l{2pt}r{2pt}){2-3} \cmidrule(l{2pt}r{2pt}){4-7}
Base architecture & AlexNet & ResNet34 & AlexNet & VGG16 & ResNet18 & ResNet34 \\ 
\midrule
\textsc{Scratch} & 25.42 & 42.72 & 26.37 & 29.78 & 39.98 & 40.82 \\
\textsc{ImageNet} & \textbf{33.44} & \textbf{50.47} & \textbf{36.27} & \textbf{49.01} & \textbf{54.04} & \textbf{56.91} \\
\midrule 
\textsc{Moving~\cite{agrawal2015learning}} & 25.57 & -- & 26.64 & -- & -- & -- \\ 
\textsc{Ego-motion~\cite{jayaraman2015learning}} & 21.89 & -- & 26.03 & -- & -- & -- \\
\textsc{Colorization~\cite{zhang2016colorful}} & 26.97 & -- & 28.25 & -- & -- & -- \\
\textsc{Context~\cite{pathak2016context}} & 25.82 & -- & 26.41 & -- & -- & -- \\
\textsc{Flow~\cite{jason2016back}} & -- & 46.09 & -- & -- & 47.95 & 50.39 \\
\textsc{Depth~\cite{zhou2017unsupervised}} & -- & 45.11 & -- & -- & 48.76 & 50.76 \\
\textsc{Depth~\cite{zhou2017unsupervised}+pose} & -- & 46.32 & -- & -- & 49.58 & 52.37 \\
\midrule 
SensorFlow & \textbf{30.48} & \textbf{49.46} & \textbf{29.35} & \textbf{36.52} & \textbf{52.97} & \textbf{54.24} \\
\bottomrule
\end{tabular}
\end{adjustbox}
{\tiny *Only \textsc{ImageNet} uses labeled data for pretraining.}
\end{minipage}
\vspace{-4mm}
\end{table*}

\subsection{Applying Learned Representation to Semantic Segmentation}

%
We examine the transferability of our learned representation to other driving tasks, by applying it to semantic segmentation in a driving environment. For this essential application in autonomous driving systems, we finetune the FCN~\cite{long2015fully} architecture and evaluate it on the CamVid and Cityscapes datasets.
Four base encoders -- AlexNet, VGG16, ResNet18 and ResNet34 -- are used for FCN.
For AlexNet, we use FCN-32s, defined in the original paper. For the VGG16, ResNet18 and ResNet34 encoders, FCN-8s is used.
Table~\ref{tab_segmenation} shows the results in terms of mean IoU for FCN with different base networks and different initialization methods, including random initialization from \textsc{Scratch}, ImageNet-pretrained model (\textsc{ImageNet}), our approach (SensorFlow), and several other unsupervised representation learning methods.
Our approach shows clear performance improvements over random initialization for AlexNet-, VGG-, and ResNet-based FCNs on both datasets, and comes close to that of supervised ImageNet in some cases, demonstrating the effectiveness of our pretrained models.


One might raise a question of whether motion information really plays an important role for representation learning. Is it insufficient to learn a representation from multiple frames using a photometric loss?
\textsc{Flow~\cite{jason2016back}} is an unsupervised optical flow learning method using a photometric loss.
Since originally it takes two concatenated frames as input, we finetuned its base network with a random initialization for the first layer, which is replaced to handle the single-image input of semantic segmentation.
The results show that learning flow through only the visual domain does not capture scene semantics while our proposed method does.
We conjecture that learning pixel displacements between images depends on local pattern matching, rather than semantic scene understanding.
In comparison, learning with motion data paired with visual domain data provides a better way for acquiring a semantic representation.

Furthermore, we compare with existing self-supervised representation learning methods that exploit ego-motion data~\cite{agrawal2015learning,jayaraman2015learning}, and that utilize appearance information such as a color or context \cite{zhang2016colorful,pathak2016context}.
Since few previous works conduct semantic segmentation as a test for representation learning, we have retrained each model, with the same experimental setup as ours. It is shown in Table~\ref{tab_segmenation} that our method yields significant improvements on the target task over both the motion- and appearance-driven methods on AlexNet.

Depth information has recently been shown to be useful for semantic tasks~\cite{jiang2018self}.
For the fair comparisons with depth-motivated representations, we validate ours with \textsc{Depth} learned on unsupervised single-image depth estimation~\cite{zhou2017unsupervised}, and the \textsc{Depth} trained with pose obtained from motion sensors, termed \textsc{Depth+pose}.
From the results, we confirm that inaccuracies in pose estimation lead to uncertainty at object boundaries. We note that while pose estimation from images is susceptible to low image quality, \eg, from adverse weather and saturated exposure, sensor data is insensitive to these factors and serves as a stable complementary modality.
Still, with given pose values, ours achieves better performance than \textsc{Depth+pose}. This may be explained by two reasons.
First, we conjecture that constraints by geometric priors, \eg, epipolar constraint, hinder learning a generic transferable representation. 
Second, reconstruction losses based on depth re-projection are known to be quite noisy, as discussed in Sec.~3.3 of Mahjourian~\etal~\cite{mahjourian2018unsupervised}. They mention that this problem could be avoided by directly learning to predict the adjacent frames.
Supported by the aforementioned results, our network is more stable to train and yields more favorable performance in comparison to existing learned representations for driving scenes.

\section{Conclusion}
In this work, we proposed a novel sensor fusion architecture that predicts a dense flow map from physical sensor readings fused with the input frame, while exploiting time symmetry for regularization.
Though our system is trained to synthesize nearby frames, the visual representation it learns can be effectively transferred to other scene understanding tasks in the driving scenario.
In particular, the transfer of our model to semantic segmentation yields leading results in comparison to existing representations acquired by unsupervised learning.

\vspace{2mm}
\paragraph{Acknowledgements} The authors gratefully acknowledge Dong-Geol Choi for assistance with data capture systems and his helpful discussions. 
This work was supported by the Technology Innovation Program (No. 10048320), funded by the Ministry of Trade, Industry \& Energy (MI, Korea).


\bibliography{egbib}
\end{document}